# The "*psychological map of the brain*", as a personal information card (file), -- a project for the student of the 21st century

*Emanuel Gluskin*

Galilean Sea Academic College and Braude Academic College, Israel,
gluskin@ee.bgu.ac.il, http://www.ee.bgu.ac.il/~gluskin/

**Abstract**: We suggest a procedure that is relevant both to electronic performance and human psychology, so that the creative logic and the *respect for human nature* appear in a good agreement. The idea is to create an electronic card containing basic information about a person's psychological behavior in order to make it possible to quickly decide about the suitability of one for another. This "psychological electronics" approach could be tested via student projects.

**The point**

The point of the following suggestion (briefly mentioned in [1]) is merely logical, not of any complicated electronic technology. Everyone might be interested in this suggestion, because it should make our life easier.

The unusual degree of freedom used in this suggestion is that the electronics (program) design is led by a preliminary *psychological test*. The union of the electronics means, making it possible to quickly read and analyze information, and the relevant psychological approach related to the individual, is seen to be constructive, promising good effects. To start with, let us note that as the human population becomes more and more dense, and close to one another *in many senses*, we often meet the uneasy necessity to well *fit* each other, usually having a very limited time for achieving this important goal. In marriages, job contacts, and some other situations, an unexpected feature of a partner or co-worker can lead to serious trouble. Just think of a girl having in her brain the center defining her love feeling being uncontrollably (unconsciously) connected with the center of anger (aggressiveness), or about a Boss's skillful driver who is unable to quietly sit and wait for half an hour, or about one whose relation to money or other "home matters" are those of a *player*, -- should a girl who wishes to create a serious family with educated children marry such a person? As well, it can be unpleasant or shocking, if your co-worker suddenly exhibits impatience with your national mentality.

Of course, one *can* accept any such disadvantage of the other, seeing the other's "pluses" as against a particular disadvantage, but it should be seen that:

a. One *has the right to know the situation a priori*, and, definitely, in such important cases as marriage or a job agreement, the information about the disadvantages should be *required* to be provided, which, as we shall explain, is physically (technically) possible.

<u>b</u>.  Even a himself-problematic person should be finally interested in not being in contact with a human environment which is not suitable for her or him.  Thus, though one does not always wish her or his psychological features to become known, it may be expected that in general people will be interested in providing (and receiving) such kinds of information; stressing that the right to obtain the information is a mutual one.

It is obviously too late, redundant, and even physically impossible, to start with the psychological investigation when a conflict already started.  However, -- when, in a friendly atmosphere organized by a psychologist, not associated with any existing conflict, just seen as a scientific or philosophical conversation, or an amusing test, one answers the questions about some relevant situations, -- a sufficiently complete picture of one's characteristics called here "connections between the main brain centers" can be easily obtained.  As the point of the technical contribution, this information is meant to be finally recorded in an electronic card (file), to be checked when it is necessary to avoid the possible dangerous conflicts as much as possible.

**The logistics of the proposition**

When speaking about the "*brain centers*", we mean, of course, some real physical (biological) centers that can, electrically or chemically, interact with each other, but not, however, the placement of the centers in the brain, nor their form (which need not even be well localized), and no such *physical* brain-maps as the known (see, e.g., Google) CT or MRI maps are required here.  We approach the brain as a "system", i.e. in the simple sense of the input-output mapping, and see the results of the *psychological* tests as a graph the vertices of which are the "main brain centers" and the branches, -- the possible connections.  While the centers are, more or less, the same for all of us, the *connections* between the centers are much more individual and they are our focus.

We thus have the main stages of the procedure as follows.

1.  A psychologist (preferably together with a sociologist) develops a test, and a person under study (presumably over 18 years old) fills a questionnaire that is then turned into a suitable electronic-file form.  Such a test should not be once per life, of course, it can be repeated (refreshed) once per several years.

2.  The informational (electronic, magnetic, etc.) card is used/involved then to include the personal data obtained in the test, this data to be easily read when it is needed.

3.  A program is developed for mutually comparing two such personal informational cards, in order to see the matching (fitting) between the card's owners and thus estimate the *probability of success* in the proposed undertaken, e.g. in marriage, -- a case when the use of the card is, perhaps, most mandatory, because stresses and scandals in families sometimes end tragically.

A draft of some such *uncompleted* psychological test is shown in Fig. 1.  The (nonzero) value of a certain coefficient $a_{ik}$ showing the probability of the influence of center number $i$ on center number $k$ is determined by some subtest, after the initial test reveals that this probability is not ignorable.  The stage of the performance of the test shown in this figure, takes place after the general part and the subtest that determines $a_{12}$ and $a_{21}$ are performed.

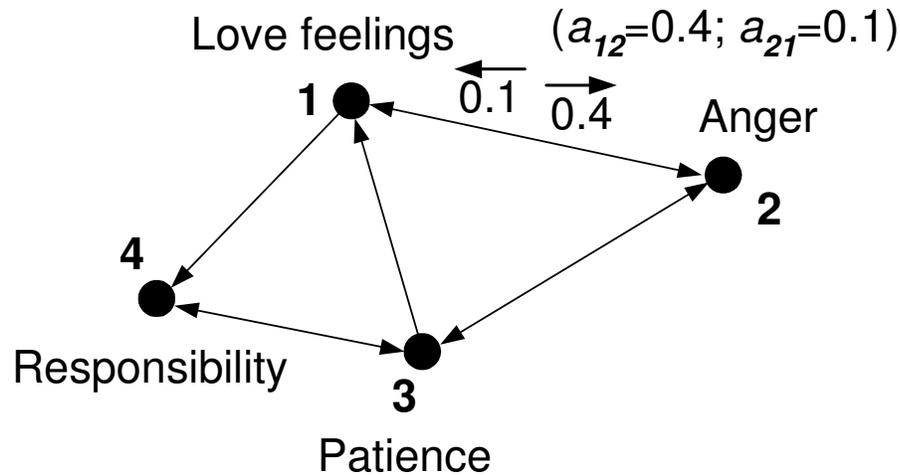

Fig 1: The psychological test of a "4-center"-person is being conducted. The *connections* of the centers are observed at first. Then the relevant subtests reveal and determine, via *intensities* of the interactions (in both directions), the *probabilities* of the influences between the centers, expressed by the numerical values of $a_{ik}$, here the already found for $a_{12}$ and $a_{21}$.

The matrix characterizing connections between the centers looks at this stage as:

$$\begin{pmatrix} a_{11} & a_{12} & a_{13} & a_{14} \\ a_{21} & a_{22} & a_{23} & a_{24} \\ a_{31} & a_{32} & a_{33} & a_{34} \\ a_{41} & a_{42} & a_{43} & a_{44} \end{pmatrix} = \begin{pmatrix} 0 & 0.4 & 0 & a_{14} \\ 0.1 & 0 & a_{23} & 0 \\ a_{31} & a_{32} & 0 & a_{34} \\ 0 & 0 & a_{43} & 0 \end{pmatrix}.$$

One agrees that the value 0.4 of the probability of influencing the center of anger of love feelings is really frightening!

   Of course, one can use the diagonal elements in order to express the relative "strengths" of the centers, having $a_{ii}$ nonzero, say, in the range $0 < a_{ii} < 1$.

 **To the students**

 The proposition of the "psychological electronics applications", which we put her forward does not belong to any well established or popular field, and it is always objectively difficult to present a new approach in a perfect form. The present general consideration is suggested, however, to be continued by the students who can invent a simple game involving some primitive creatures (or a "society" of these creatures), having very few of the "brain centers".

   Take for the first experiment 2 centers (e.g. those of *hunger* and *justice*) which can be mutually connected with some *individual* $a_{12}$ and $a_{21}$. Give some distributions for $a_{12}$ and $a_{21}$. What is the state of mood (defined by you as some quantity) of this funny "society"? Are your creatures "happy"? Deriving some conclusions, try to pass to a model of 3 centers, etc..

The competition of such games (projects) might be an interesting event in the school or the college.

**The extended card**

One can also propose an *extended information card* that would include, in addition to the information about brain centers, some important information about the person, provided by those who know him well, and on whom he can rely.

It is quite reasonable to ask, e.g., your old mom (who hardly will be able to help you with her advice in some 20 years), or your older friend who well understands you, to add to the card their advice as regarding what would suit you well; to what kind of people you should be (or not be) close. The fact is, that under many circumstances, when we are attracted by something, we forget about the great treasure given to us by society, -- the wise advice of those who are close to us, love us and can correctly see us "from outside". Thus, recording some good advice for future reference can be important. In particular, you can be sure that your girl will be very interested in the advice and the characteristic ascribed to you by your mother.

For instance, one's mother might say: "*He will always choose the most difficult way*", and one's older friend might say: "*Marriage to a religious girl would be a stable state for him.*"

Even you yourself can quietly "observe/consider yourself" and record on the card some important advice for you future.

Thus, assume that you think about yourself: "*I always get more important results when I think about something simple, than when I try to quickly reach the heights*" and: "*I should always count up to 10 before responding*".

You can derive from these two thoughts the brief: "*It is desirable for me to be in a quite work atmosphere*" which is constructive for deciding whether or not to work with another person, or in a certain human environment, and worth recording in some technically proper form.

Perhaps, such free-will addition to the information card can be done via some special entrances provided in an extended psychological test. In any case, the target of presenting the information in an electronic form for automatic comparison of different cards becomes more difficult with the extension and finding a common electronic format for the "*Brain Centers*" and the "*Friendly Advices*" parts of the collected information would be an interesting solution.

One sees that some freedom exists already in the basic logic of the initial psychology test, and in the way of presenting and comparing the information.

Last, but not least, since technological developments have their own laws and tendencies which by themselves have no connection to human moral, the involvement, via the psychological aspect, of the **respect for human nature**, may be a point for correct direction of the future applied science.